\documentclass{article}
\usepackage[margin=1in, left=1.5in, right=1.5in,]{geometry}

\usepackage{graphicx}
\usepackage{xcolor}
\usepackage[noblocks]{authblk}

\usepackage[square,numbers]{natbib}
\usepackage{hyperref}
\usepackage[noabbrev]{cleveref}

\hypersetup{
    colorlinks,
    linkcolor={red!50!black},
    citecolor={blue!50!black},
    urlcolor={blue!80!black}
}

\usepackage{url}

\makeatletter
\renewcommand{\maketitle}{\bgroup\setlength{\parindent}{0pt}
\begin{flushleft}
  {\LARGE{\textbf{\@title}}} \\
  \vspace{2ex}
  {\large \@author}
\end{flushleft}\egroup
}
\makeatother

\renewcommand{\cite}{\citep}

\title{}

\begin{document}

{\centering \LARGE
\textbf{Reflections from the Workshop on \\ AI-Assisted Decision Making for Conservation \\}
}
\vspace{3ex}

\author[1,*]{Lily~Xu}
\author[2,3,4,*]{Esther~Rolf}
\author[5]{Sara~Beery}
\author[6]{Joseph~R.~Bennett}
\author[7,8,9,10]{Tanya~Berger-Wolf}
\author[11,$^\dagger$]{Tanya~Birch}
\author[5,12]{Elizabeth~Bondi-Kelly}
\author[13]{Justin~Brashares}
\author[13]{Melissa~Chapman}
\author[14]{Anthony~Corso}
\author[15]{Andrew~Davies}
\author[16]{Nikhil~Garg}
\author[17]{Angela~Gaylard}
\author[18]{Robert~Heilmayr}
\author[19]{Hannah~Kerner}
\author[20]{Konstantin~Klemmer}
\author[21]{Vipin~Kumar}
\author[20]{Lester~Mackey}
\author[4,22]{Claire~Monteleoni}
\author[15]{Paul~Moorcroft}
\author[23]{Jonathan~Palmer}
\author[7]{Andrew~Perrault}
\author[24]{David~Thau}
\author[1,2,11]{Milind~Tambe}

\affil[1]{School of Engineering and Applied Sciences, Harvard University}
\affil[2]{Center for Research on Computation and Society (CRCS), Harvard University}
\affil[3]{Harvard Data Science Initiative (HDSI)}
\affil[4]{Department of Computer Science, University of Colorado Boulder}
\affil[5]{Electrical Engineering and Computer Science, Massachusetts Institute of Technology}
\affil[6]{Department of Biology, Carleton University}
\affil[7]{Department of Computer Science and Engineering, The Ohio State University}
\affil[8]{Department of Electrical and Computer Engineering, The Ohio State University}
\affil[9]{Department of Evolution, Ecology, and Organismal Biology, The Ohio State University}
\affil[10]{Translational Data Analytics Institute, The Ohio State University}
\affil[11]{Google}
\affil[12]{Electrical Engineering and Computer Science, University of Michigan}
\affil[13]{Department of Environmental Science, Policy, and Management, University of California Berkeley}
\affil[14]{Department of Aeronautics and Astronautics, Stanford University}
\affil[15]{Department of Organismic and Evolutionary Biology, Harvard University}
\affil[16]{Operations Research and Information Engineering, Cornell Tech}
\affil[17]{African Parks}
\affil[18]{Bren School of Environmental Science and Management, University of California, Santa Barbara}
\affil[19]{School of Computing and Augmented Intelligence, Arizona State University}
\affil[20]{Microsoft Research}
\affil[21]{Department of Computer Science and Engineering, University of Minnesota}
\affil[22]{INRIA Paris}
\affil[23]{Wildlife Conservation Society (WCS)}
\affil[24]{WWF Global Science}

\date{}


\begin{abstract}
\normalsize    In this white paper, we synthesize key points made during presentations and discussions from the AI-Assisted Decision Making for Conservation workshop, hosted by the Center for Research on Computation and Society at Harvard University on October 20--21, 2022. We identify key open research questions in resource allocation, planning, and interventions for biodiversity conservation, highlighting conservation challenges that not only require AI solutions, but also require novel methodological advances. In addition to providing a summary of the workshop talks and discussions, we hope this document serves as a call-to-action to orient the expansion of algorithmic decision-making approaches to prioritize real-world conservation challenges, through collaborative efforts of ecologists, conservation decision-makers, and AI researchers.
\end{abstract}

\maketitle
\thispagestyle{empty} 

\noindent {\fontsize{9}{9}\selectfont * These authors contributed equally}

\noindent {\fontsize{9}{9}\selectfont $^\dagger$ The opinions are the authors' and don't necessarily reflect those of Google.}

\clearpage

\section{Introduction}
\label{sec:intro}
 
Effective area-based conservation management is more imperative than ever before \cite{cardinale2012biodiversity,ceballos2015accelerated,jones2018one,watson2021talk}. The world has lost one-third of global forests \cite{williams2003deforesting} and one million species face extinction due to human activities \cite{tollefson2019humans}. To reverse these trends, conservation decision-makers are tasked with managing protected areas \cite{watson2014performance,coad2015measuring}, prioritizing endangered species \cite{chades2008stop,schuster2019optimizing,ball2009marxan}, and navigating opportunity and cost trade-offs between conservation and economic development \cite{fox2019efficient,sacre2020relative,dowd2022economic} --- often in the face of uncertainty, complex constraints, and long planning horizons \cite{visconti2010conservation,carvalho2017spatial,pressey2013plan,tulloch2013incorporating}. Simultaneously, recent years have seen rapid advances in artificial intelligence (AI), including deep learning to uncover patterns in high-dimensional datasets \cite{lecun2015deep}, tractable approaches to navigate exponentially large problem spaces \cite{browne2012survey}, and statistical inference to learn causal effects of actions \cite{pearl2009causal}. 

The majority of conservation areas around the world lack funding, personnel, and equipment \cite{coad2019widespread}. Without an increase in the resources available for conservation areas, it is critical to prioritize and scale the impact of the scarce resources we have. Much existing work at the intersection of AI and conservation focuses on extracting basic information from primary data, largely in the form of computer vision for biodiversity monitoring, such as supervised learning approaches to identify animals or classify land cover \cite{tuia2022perspectives,weinstein2018computer,qin2022review}. 
In addition to these applications, AI holds potential to contribute to \emph{prioritization} of resources within protected areas and to improve the execution of conservation area \emph{management} activities at scale, for example through optimization, sequential planning, and causal inference.

To this end, the AI-Assisted Decision Making for Conservation workshop convened to explore opportunities to expand the breadth of AI methods --- particularly for optimization, planning, and evaluation --- that are being developed for and applied to conservation. In this report, we summarize the discussions held and challenges raised during the workshop surrounding conservation area management; the role of AI, Earth observation, and causality in conservation decision making; and limitations in the application of AI for conservation. 

\begin{figure}
    \centering
    \includegraphics[width=\textwidth]{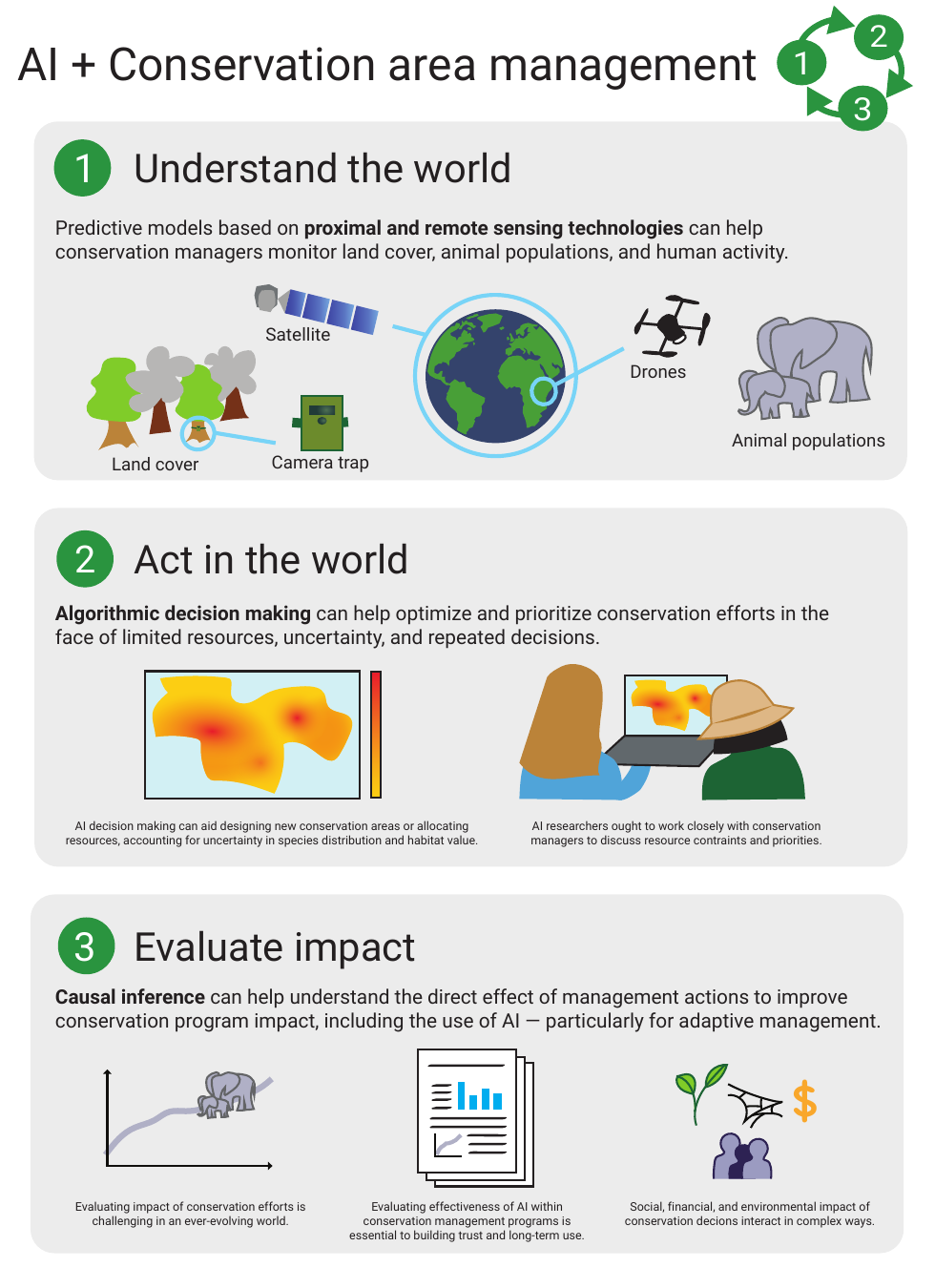}
    \caption{Effective conservation management is a planning, implementation, and adaptation cycle consisting of three key steps. First, we must \emph{understand the world}; proximal and remote sensing technologies can help monitor land cover and animal populations (\cref{sec:earth-observation}). Second, we must \emph{act in the world}; algorithmic decision making can help optimize and prioritize conservation efforts (\cref{sec:ai-decision-making}). Third, we must \emph{evaluate the impact} of our actions; causal inference can help understand the direct effect of our actions (\cref{sec:causality}).}
    \label{fig:conservation-management}
\end{figure}

First, we synthesize key conservation priorities. For effective conservation area management, the primary steps are to:
\begin{enumerate}
\item \textbf{Understand the world.} To inform conservation decision making, managers must first develop an understanding of the status of animal populations, species distributions \cite{guisan2013predicting,beale2012incorporating}, land cover \cite{bailey2016land}, animal threats \cite{kuiper2022robust}, and human activity \cite{goudie2018human}. 
\item \textbf{Act in the world.} Using their best understanding of the world, managers are tasked with taking conservation actions like designating new conservation areas \cite{ball2009marxan}, allocating ranger patrols \cite{critchlow2017improving}, implementing human–wildlife conflict deterrence programs \cite{kuiper2022robust}, intervening in the illegal wildlife trade \cite{gore2023advancing}, securing sustainable harvest \cite{memarzadeh2019rebuilding}, supporting sustainable economies \cite{teytelboym2019natural}, and countless other decisions --- almost always in the face of limited resources and uncertainty.
\item \textbf{Evaluate impact.} Managers want to evaluate the outcomes of their actions in order to learn whether certain actions are achieving desired impact \cite{jayachandran2017cash}, and to iteratively improve decision making within complex, interactive systems \cite{levin2013social,wilson2021causal}.  
\end{enumerate}
We discuss these steps in detail in \cref{sec:conservation-management}; see \cref{fig:conservation-management} for a visual summary.

AI methods have the potential to aid in each of these three components of conservation management. Some approaches have already been in use for years; other approaches are not yet deployable but are growing in promise. Algorithmic approaches can help:
\begin{enumerate}
    \item \textbf{Understand the world.} AI for Earth observation can help across multiple scales and modalities, from studying changes using remote sensing imagery \cite{brodrick2019uncovering,lobell2015scalable,hansen2013high,rolf2021generalizable,khandelwal2022realsat,brown2022dynamic,reichstein2019deep,kellenberger2019few} to monitoring individual animal movement with photographs \cite{lahiri2011biometric,pitman2017cats}, acoustic monitoring \cite{sugai2019terrestrial}, and GPS tracking \cite{browning2018predicting}. This is the focus of \cref{sec:earth-observation}.
    \item \textbf{Act in the world.} AI methods can help plan across a range of contexts relevant to conservation, including sequential settings, under model uncertainty, and in the presence of multiple agents \cite{kochenderfer2022algorithms}. This is the focus of \cref{sec:ai-decision-making}.
    \item \textbf{Evaluate impact.}  One important aspect of evaluating the impact of conservation decisions is isolating causal effects of these decisions. Causal inference techniques can help understand the \emph{causal} effect of conservation actions, including scenarios where randomized controlled trials are too expensive or infeasible to conduct \cite{beyers1998causal,ferraro2019causal,butsic2017quasi}. In addition, understanding causal relationships between elements of the environment, such as the effect of elephant populations on carbon stocks in savannas \cite{davies2019elephants}, can aid in designing comprehensive management decisions.
    This is the focus of \cref{sec:causality}.
\end{enumerate}

\begin{figure}[h]
    \centering
    \includegraphics[width=.8\textwidth]{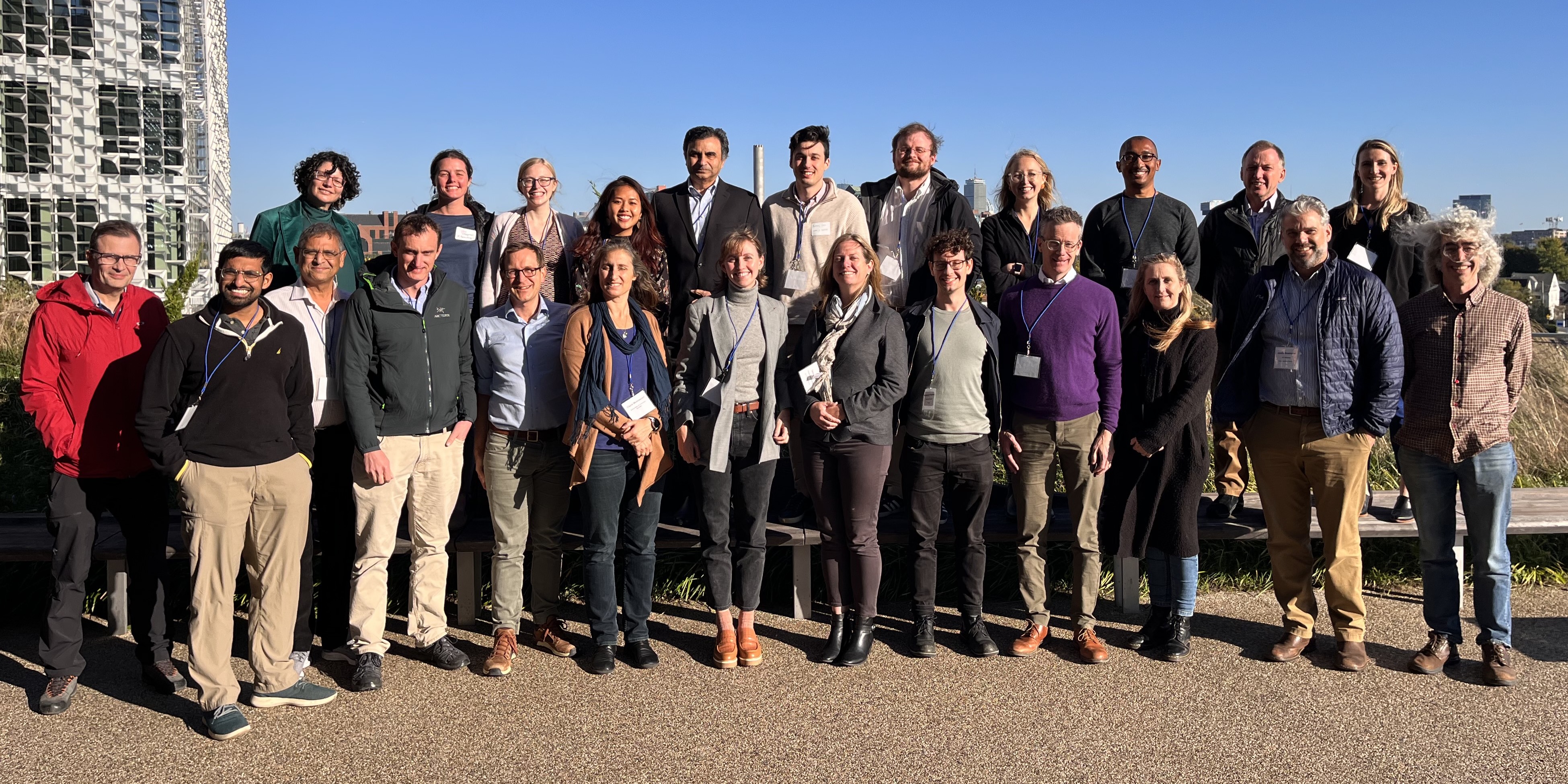}
    \caption{The AI-Assisted Decision Making for Conservation workshop was hosted by the Center for Research on Computation Society (CRCS) at Harvard University in October 2022. The co-authors of this paper were the workshop participants, coming from academia, industry, and non-profits with expertise in computation, conservation, and deploying research.}
    \label{fig:group-photo}
\end{figure}

In this white paper, we recollect and synthesize key points made during presentations and discussions from the 2022 AI-Assisted Decision Making for Conservation workshop (\cref{fig:group-photo}). Where possible, we identify key open research questions in resource allocation, planning, and interventions for biodiversity conservation that arose in whole-group or partial-group conversations. In particular, we pinpoint conservation challenges that not only require AI solutions, but require novel methodological advances. While this synthesis document is not meant to serve as an entirely complete or comprehensive research agenda, we nonetheless hope that it can help \textbf{orient the expansion of algorithmic decision-making approaches so that they prioritize real-world conservation challenges}, through collaborative efforts of ecologists, conservation decision-makers, and AI researchers.    

\vspace{6ex}

\vspace{4ex}

\noindent\textit{A full schedule of the two-day workshop, with PDF slides from all talks, is available on the workshop webpage: \url{https://crcs.seas.harvard.edu/conservation-workshop}.}

\vspace{4ex}
\clearpage
\section{Primer to conservation area management}
\label{sec:conservation-management}

Conservation areas are defined geographical spaces on land or sea managed for biodiversity and climate mitigation. These include protected areas \cite{watson2014performance}, Indigenous and Community Conserved Areas (ICCAs) \cite{berkes2009community}, and Other Effective area-based Conservation Measures (OECMs) \cite{dudley2018essential}.
Initiatives like 30x30, which aims to protect 30\% of land and sea by 2030, underscore the importance of protected areas for the sustainable development goals (SDGs) and for the future of life on Earth \cite{dinerstein2019global,maxwell2020area}. 

For the scope of this document, we focus on field conservation, namely conservation area management and species management. Field conservation efforts include preventing human-wildlife conflict \cite{wilkinson2020ecological}, processing multi-modal inputs from sensors including camera traps \cite{marvin2016integrating,thau2019artificial,ahumada2020wildlife,speaker2022global}, supporting Indigenous and local communities \cite{schuster2019vertebrate,gore2016local}, supporting women \cite{agu2020women}, and tracking and predicting animal movement \cite{smouse2010stochastic}. The full scope of conservation initiatives is much broader, as documented by the Conservation Measures Partnership \cite{dietz2010increasing,redford2018assessment}. Other activities include raising awareness, providing alternative livelihoods \cite{cooney2017poachers}, and creating policy guidelines \cite{emogor2021scale}. 

Ecosystems are highly complex dynamic systems, making conservation planning challenging. If we don't know the answer to simple questions of monitoring, such as:  \textit{what is the range of the western gorilla?}, then it is difficult to answer a host of more complex yet critically important questions such as: \textit{how do we spatially prioritize land to protect? Does deforestation lead to increased human-wildlife conflict? Does a conservation program to protect elephants improve people's livelihoods?}

In this section, we provide an overview of management theory and current conservation practices while highlighting key challenges in monitoring, decision making, and impact evaluation. 

\paragraph{Prioritization} Conservation prioritization includes both deciding how best to allocate conservation resources in areas that are already protected as well as deciding where to expand existing conservation measures. The designation of new protected areas is part of the planning process, which requires taking into account the locations of existing protected areas, landscape connectivity, and biodiversity distributions. 

\paragraph{Adaptive management} Adaptive management is a planning, implementation, and adaptation cycle \cite{gaylard2011advances,bennett2015polar}. The goal of adaptive management is to manage ecosystems using the best available information, accounting for feedback between decisions and outcomes. 

Adaptive management approaches are complicated by exogenous changes, including climate change \cite{kaack2022aligning}, which require repeated cycles of monitoring, planning, and action. Often, \emph{sequences} of events are important: for example, availability of surface water impacts elephant movements, but also depends on previous years' rainfall.  Other sources for dynamic change include species migration, seasonal weather patterns, and species interaction (including predator--prey interaction and habitat competition). 

\paragraph{Evaluation and evidence-based management} Adaptive management is closely related to evidence-based management \cite{hayes2019designing}, which centers around using repeated assessments to support decision making. Defining metrics of conservation assessment --- what counts as conserved, and/or what defines successful protected area management --- is critical to ensuring conservation strategies contribute to underlying objectives of protected areas.  For example, land area being classified as protected does not necessarily mean it is well-resourced or effectively managed. The metrics we define to evaluate the outcomes of protected area management, such as changes in population abundance or land use, can then be used to update the management strategy (adaptive management).

\vspace{4ex}

\begin{figure}
    \centering
    \includegraphics[width=.8\textwidth]{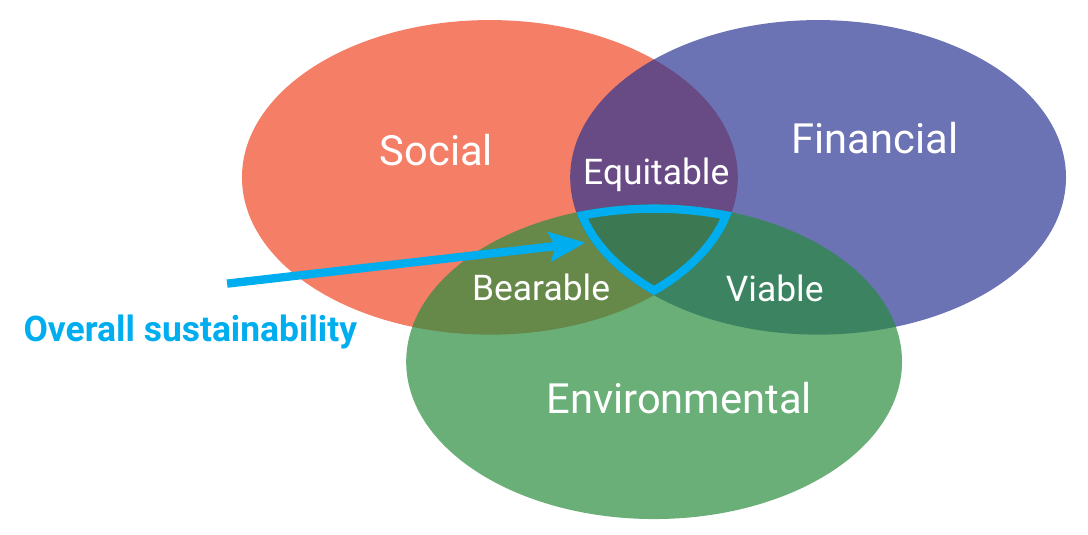}
    \caption{Overall sustainability must satisfy social, environmental, and financial considerations. These solutions ought to be viable to implement, equitable to impacted communities, and bearable to both humans and ecosystems. 
    (Figure adopted from the triple-bottom line sustainability framework \cite{elkington1998accounting}.)
    }
    \label{fig:sustainability-venn}
\end{figure}

In addressing these challenges, we need to think about simultaneously optimizing for environmental, social, and economic sustainability (see \cref{fig:sustainability-venn}). Social and economic benefits and risks must be considered in tandem with environmental impacts \cite{bennett2017conservation,adams2004biodiversity}. For example, increased lion conservation near human habitation may come with risks to those neighboring communities by increasing the risk of human–wildlife conflict \cite{treves2006co}. 
Accounting for the many facets of sustainability can be an effort stemming from many communities working together \cite{brashares2004bushmeat}.
The move towards truly participatory planning requires community engagement to empower Indigenous and local communities and integrate their expertise \cite{shepard2002advancing,vitos2013making,bondi2021envisioning}. For computational scientists, this includes reducing barriers to access to conservation tools --- including Wildlife Insights \cite{ahumada2020wildlife}, eBird \cite{sullivan2009ebird}, iNaturalist \cite{van2018inaturalist}, SMART \cite{rainey2021integrated}, and decision-facilitation tools.

\clearpage

\section{AI can assist conservation decision making}

AI has the potential to aid in different stages of the conservation management cycle through data-driven decision support.
In conservation area management, AI is already used to aid at every step of setting goals, monitoring resources (the step at which AI is most commonly involved), taking management action and measuring outcomes (some AI methods at present), analyzing trends and refining the approach (some AI methods at present) \cite{fang2019artificial,kwok2019ai,speaker2022global}. 

For monitoring, AI can be paired with Earth observation data to help understand the world. Examples include predicting areas of deforestation risk \cite{mayfield2020considerations}, estimating carbon stock \cite{reiersen2022reforestree}, and learning and downsizing climate variables \cite{scoville2021algorithmic,climalign2021}. AI can help extract relevant scientific information from passive and active sensor data, including images (satellites, camera traps, citizen science platforms, drones), acoustic data, motion tracking, on-body (e.g., radio collars), in-situ sensors, eDNA measurements, and even media sources, including social media, for alerts of environmental hazards and to monitor public sentiment of deployed programs. AI can also help overcome biases in data collection and missing data \cite{dobson2020making}.

For decision making, AI can be used for resource allocation (e.g., patrols, sensor deployment, and prioritization), strategic reasoning, and multiobjective optimization. AI can also be a tool for increasing equity around language variation via internalization and auto-translation. AI can also enable participatory action, both through intelligent agents and by making humans more efficient and able to scale.

Lastly, for evaluating impact of actions, combining causal inference with AI has the potential to describe trends in conservation.

At the same time that AI can be powerful in reducing barriers and access to data for protected area managers and partnering communities, it can inadvertently serve to reinforce existing power structures. We discuss the interconnectedness and complexity of design goals and system impacts in \cref{sec:critical-considerations}.

\subsection{Earth observation: AI to help understand the world}
\label{sec:earth-observation}

The immense scale at which many conservation decisions are made requires comprehensive monitoring of the Earth over time and space. AI techniques can extract higher-order information from datasets of raw observations collected at different spatial and temporal resolutions across the globe. The outputs from AI models, such as machine learning predictions, may be useful for conservation decision making directly or may be combined as inputs to other downstream analyses to produce information needed for decision making.

\begin{figure}
    \centering
    \includegraphics[width=\textwidth]{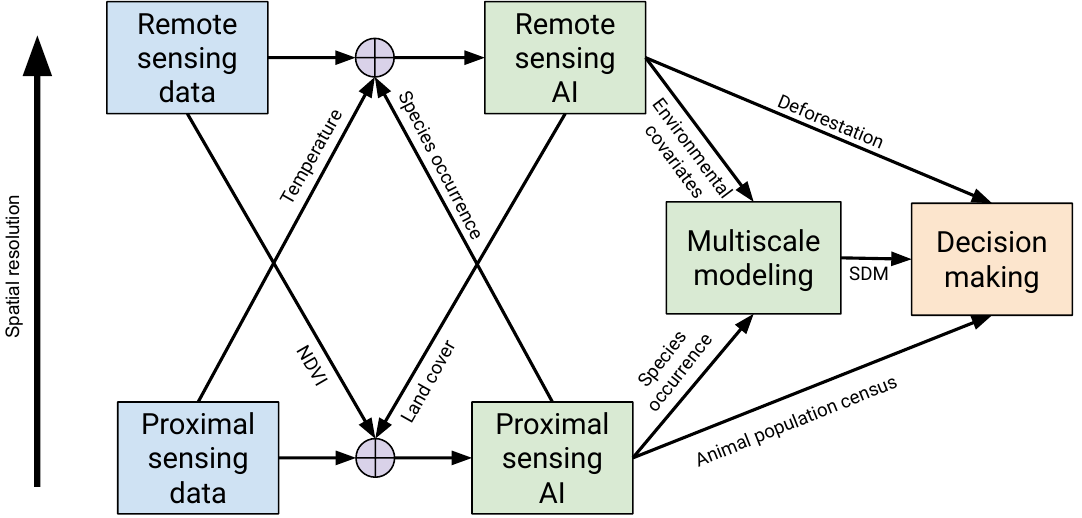}
    \caption{An illustrative example of how proximal and remote sensing approaches, the processing of the captured data, and AI can be integrated to inform decision making. The vertical axis roughly represents the spatial scale of the observations (from most proximal to most remote), the horizontal axis roughly represents the steps of the decision process, from data sourcing via modeling to decision making. Purple circles denote data (dis)aggregation and (pre)processing steps.}
    \label{fig:proximal-remote-sensing}
\end{figure}

The field of Earth observation involves the collection of physical, chemical, and biological data about planet Earth and its surface, often using sensors with variable resolution, sampling, and instrumentation from the ground up to space. \Cref{fig:proximal-remote-sensing} shows an illustrative example of how AI methods can extract information from raw data to aid conservation management decisions.

\subsubsection{Data collection and characteristics}

We distinguish Earth observation sensors into two main categories: proximal sensing (e.g., field data from camera traps, acoustic monitors, temperature sensors) and remote sensing (i.e., satellites, UAVs). 

Proximal data are collected ``in situ'' and close to the target of interest, using on-the-ground sensors. Throughout the last decade, the emergence of affordable IoT devices and robots has enabled the creation of vast sensor networks with pervasive spatial coverage, such as networks of camera traps for monitoring wildlife distribution or urban air quality sensors. The appropriate spatial and temporal allocation of sensors is a central determinant of the success of proximal sensing. 

Remotely sensed data are captured using sensors that are distant from the target of interest, such as satellites. More pervasive than ever, satellites and other sensors provide near real-time data streams of optical imagery (RGB and multispectral), synthetic-aperture radar (SAR), and more. Remotely sensed data are characterized by their tremendous spatio-temporal heterogeneity, which leads to distinct modeling challenges. For example, resolution, frequency, and quality of observations may vary across time and space depending on sensor availability. In addition, specific spatial, temporal, and spectral resolutions (e.g., the resolution of a satellite camera, the fly-over frequency, or the measured wavelengths) may constrain modeling efforts with remotely sensed data. 

For enabling evidence-based conservation management, proximal and remotely sensed data can be valuable individually or in combination. 
The proximal vs. remote sensing distinction is an oversimplification reflecting  end-points of a range of s spatial scales (e.g., sensors onboard drones and aircraft lie somewhere between proximal and remote sensing). Likewise, AI methods are not constrained to working with either remote or proximal sensing data individually, but can also combine multiple scales into one modeling framework. 

Data from different sources can be complementary \cite{moore2020comparison}: a camera trap (proximal sensor) might take a picture of a buffalo on the ground, while a satellite (remote sensor) with a high-resolution camera might take an overhead picture of the same animal. The two sources can be combined for a more comprehensive buffalo  census than either source independently would allow. 

\subsubsection{Data analysis and modeling}

Machine learning approaches, particularly computer vision, have challenged traditional image processing techniques in the analysis of remotely sensed data. The large scale and pervasiveness of the available data can synergize well with the ``data-hungriness'' of machine learning. However, labeled data are often scarce, especially in conservation-related domains. This scarcity of labeled data motivates dedicated learning approaches that are (1)~resource efficient (e.g., zero- or few-shot learners \cite{sumbul2017fine}) and (2)~able to learn in an unsupervised or weakly-supervised manner. Spatio-temporal dependencies in the data are another source of complexity 
 \cite{meyer2022machine, rolf2023evaluation}. 

Machine learning methods for Earth observation data have begun to adapt to these particular challenges \cite{reichstein2019deep}. These developments are evidenced by emerging domains in methodological machine learning research, such as physics-informed deep learning \cite{wang2020towards}, spatially-explicit learning \cite{klemmer2022spate,tseng2023lightweight}, weakly-supervised learning using spatial data \cite{brown2022dynamic,variational2021sinha,rolf2022resolving} or spatial representation learning \cite{yin2021gps2vec}. 

As such, dedicated methodological research is also required to generalize and adapt existing computer vision approaches to --- or develop new customized approaches for --- the remote sensing domain in use for conservation. For example, computer vision algorithms for natural images can take advantage of the abstractions of ``foreground'' and ``background,'' but there is not necessarily a notion of foreground or background in remote sensing images. Object detection tasks can also differ substantially in conservation images (like satellite or aerial imagery) compared to traditional computer vision for natural images, due to the existence of a large number possible objects across a variety of spatial scales.

\subsubsection{Open challenges in AI for Earth observation}

Concurrent revolutions in data (due to new satellite missions and other sensor data availability), compute (due to cloud infrastructure and cloud-hosted open datasets), and analysis methods have enabled tremendous advances in the development and application of AI for remote sensing and Earth observation. At the same time, these advances have exposed several challenges to be addressed. Here we outline several challenges in AI for Earth observation that are of key concern in conservation and decision-making settings.

\paragraph{Data availability and transparency}  The open availability of remote sensing imagery (Landsat, Sentinel, Planet/NICFI) has spurred the creation of derivative datasets (e.g., Global Forest Change, Dynamic World). At the same time, there are not always direct incentives for releasing source training data or validation data, or the models used to create the derivatives. In addition to availability, reporting the accuracy and applicability of measured or predicted datasets is particularly important \cite{meyer2022machine,gebru2021datasheets}: errors in ground-truth labels or features may propagate throughout model training and into deployment, and any downstream policies or analyses. Even in cases where ground truth data are less sparse, human errors can persist, requiring substantial data pre-processing to be suitable for model training \cite{brown2022dynamic}. 

\paragraph{Model interpretability and transparency} As dynamic data streams are constantly changing, models --- and the ways people use them --- need to be dynamic as well. This introduces additional challenges for model interpretability, transparency, and explainability, toward answering questions such as: \textit{When should the model be used? Who is the intended user? What is the model intended to do? What are the challenges and risks of using the model outputs? If the models themselves are changing constantly, does that make them usable in unseen environments or different spatial-temporal contexts?}

\paragraph{Missing labels} While many labeled datasets have been published in recent years for training and evaluating AI models, existing labeled datasets for remote sensing are relatively small and sparse. This is in part because labels for Earth observation data can be difficult to annotate, for example due to overlapping ecotones between classes in remote sensing inputs. There is often a need for domain expertise for annotation, and many use cases require ground-truthing (e.g., on-the-ground observation) to determine labels. 

\paragraph{Label biases} Furthermore, both the sampling and authoring institutions of existing labeled datasets are mostly biased toward North America and Europe \citep{luccioni2022bugs}, which in turn results in disparate levels of attention and model performance across geographies (a phenomenon that has also been observed for traditional AI data modalities \cite{shankar2017no}). This challenge is particularly  important in conservation, as ecological systems vary substantially throughout different geographical areas and models trained on data-rich environments in the Global North might not be transferable to the Global South \cite{zhang2022segmenting}. Future efforts could investigate how to more efficiently, scalably, and equitably collect labeled datasets for remote sensing and AI. 

\paragraph{Generalization} Generalization of models across spatial (e.g., from one landscape to another), temporal (e.g., from rainy season to dry season), and spectral dimensions (e.g., from RGB to thermal imagery) is an important open challenge for conservation that may be aided by methods for transfer learning or distribution shift. 

\paragraph{Responsibility and governance} While the increased availability of data, methods, and compute for AI and remote sensing is  empowering for many researchers and practitioners, it also raises important ethical concerns. \textit{Who is responsible for the output of a model (predicted map) and how it gets used? How do we prevent bad actors from misusing data and methods? Who has the authority to make maps, particularly given the history of extraction (local ownership, parachute science \cite{asase2022replacing}, and data colonialism)? How can a predicted map be used measure the impact of efforts on conservation or sustainable development goals?}

\paragraph{Increased collaboration} The AI, Earth observation, and conservation research communities can benefit from more direct collaboration across fields \cite{nakalembe2023considerations}. For example, collaboration between the satellite remote sensing community and the biodiversity monitoring community (camera trap, acoustic, eDNA \cite{mena2021environmental}) could lead to sharing of methods as well as development of new multi-modal methods levering the breadth of available data.

\paragraph{Computational costs} The size (and constant growth) of both proximal and remotely sensed datasets also poses computational challenges to store, process, and analyze. The scale of many remote sensing datasets means that traditional methods can be computationally expensive; AI systems may provide faster modeling, particularly at inference time, but large-scale geospatial extrapolations can still be computationally intensive, even (or especially) using the latest ML models. As such, the development of large-scale cloud compute infrastructure over recent decades has benefited the Earth observation community substantially. We discuss computational costs in greater detail in \cref{sec:critical-considerations}.

\subsection{Algorithmic decision aids: AI for decision making}
\label{sec:ai-decision-making}

\subsubsection{The role of AI in adaptive management}
As discussed in \cref{sec:conservation-management}, adaptive management is a cycle consisting of planning, implementing, and adapting conservation management decisions. AI can provide a robust means of integrating evidence into adaptive management and has the potential to deal with complexity in ways that spatial statistics, simple heuristics, or general rules of thumb cannot. Management decisions in conservation are often complex, with trade-offs among multiple objectives \cite{xu2022ranked} and limited information as a basis for decisions \cite{game2014conservation}. Such complexities amplify when the impacts of decisions must be predicted across large spatial and temporal scales, for example when planning for migratory bird conservation \cite{schuster2019optimizing}, building climate change resilience into protected area networks \cite{alagador2014shifting}, and accounting for socio-political risk in conservation planning \cite{schuster2022protected}. 

\subsubsection{Sequential decision making}
We can define a decision problem by an objective, a set of possible actions over time, and a model of the effects of those actions in the environment. Many decision problems in biodiversity conservation involve reasoning about a sequence of decisions for settings with complex dynamics and large amounts of uncertainty. AI methods can be applied to build models of the system dynamics, infer unobserved variables from observations, and to plan with those models and observations to determine an optimal sequence of actions under uncertainty \cite{xu2021dual,kochenderfer2022algorithms}. Examples discussed over the course of the workshop included:
\begin{itemize}
\item \textbf{Patrolling to interdict poachers} \cite{fang2016deploying,xu2020stay}: The objective is to deter poachers by deciding when and where to send rangers within a protected area. Poaching activity is only observed indirectly through the discovery of traps and encampments, and only in regions where the rangers patrol. The behavior of poachers and their response to changes in patrolling behavior poses a challenging modeling task, which may require robust planning \cite{xu2021robust}.
\item \textbf{Reintroduction of species} \cite{sanderson2002planning}: The objective is to ensure the long-term viability of the species by deciding when, where, and how many individuals should be reintroduced. The effects of reintroduction require ecological models, which can be informed by tracking the locations and movements of the individuals via collars or sensing modalities.  
\item \textbf{Protected area design} \cite{hanson2017prioritizr, silvestro2022improving}: The objective in protected area design is to increase species richness, preserve threatened species, or generally improve the connectedness and resilience of protected areas, while satisfying constraints of cost or total protected area. Decisions include where and when to purchase land, and making purchases now or delaying in order to gather more information or wait for opportunities. We have many tools to observe the land attributes we care about, but our observations are highly imperfect. Sometimes, lands are are entirely inaccessible for field surveys. Time horizons for protected area design are often very long and ideally evaluation is done against a wide range of possible scenarios, including budgetary uncertainty.
\end{itemize}

Sequential decision problems can be modeled as Markov decision processes (MDPs), which account for uncertainty in the dynamics, and partially-observable Markov decision processes (POMDPs), which account for uncertainty in the state of the world. Recently, AI has yielded rapid advances in methods for solving large and complex MDPs and POMDPs such as deep reinforcement learning \cite{arulkumaran2017brief} and belief state planning \cite{sunberg2018online}. These advances offer the potential to rigorously optimize sequential decisions for conservation \cite{nicol2013adaptive,chades2021primer,lapeyrolerie2022deep}. The parallel advances in machine learning–enabled sensing methods (see \cref{sec:earth-observation}) accelerate this potential by providing data that can be used to create dynamic models and infer the system state, which in turn can inform more precisely targeted and useful actions.

\subsubsection{Open challenges in sequential decision making for conservation}

\paragraph{Imperfect or unknown models} Previous successes of algorithmic decision making have been concentrated in domains that can be accurately represented by a known model, such as games \cite{silver2018general}, transportation systems \cite{kochenderfer2012next}, or subsurface applications \cite{wang2022sequential}. However, in conservation, a key challenge to applying algorithmic decision making is building accurate models of the environment dynamics and the effect of actions. Data-driven approaches have been successful at modeling complex dynamical systems, but often require large amounts of data or additional expert knowledge in order to generalize well. Since no model is perfect, model uncertainty should be accurately accounted for so that plans derived from the model work well in environments that deviate from the model. 

\paragraph{Challenging optimization} Finding an optimal strategy in well-described environments may also be difficult. Decision problems in conservation often involve long time horizons or high-dimensional and continuous state, action, and observation spaces. Modern solvers have begun to scale to such problems \cite{sunberg2018online}, but more work needs to be done to improve these optimization techniques. 

\paragraph{Integrating human expertise} In addition to pure algorithmic development, approaches that integrate expert human knowledge in the planning loop may also be needed.

\paragraph{Mismatch between predictions and decisions} Conservation decision making often involves machine learning as a substep to generate predictions, such as predicting species populations or identifying areas vulnerable to illegal mining. But then those predictions are ultimately used for some downstream decision, such as selecting areas to conserve as species habitat or add monitoring stations. Thus, aligning predictions with decision making is another priority \cite{donti2017task}.

\paragraph{Validation} Once a plan has been generated, it is crucial to validate that plan for efficacy and safety \cite{corso2021survey}. Often, high quality plans can be developed from simple models, but we need extensive validation to check that the plan is indeed of high quality. Validation can be done by building additional validation models \cite{kochenderfer2012next}, stress-testing the plan to discover the most likely failure modes \cite{lee2020adaptive}, and testing the plan on a small scale before larger scale deployment \cite{kochenderfer2012next}. That said, no model accounts for all possible consequences of actions, so additional validation is needed to anticipate unforeseen consequences of the proposed actions, such as negatively affecting marginalized groups, or damaging other aspects of the environment.

\paragraph{Unknown unknowns} A final challenge in sequential decision making involves so-called ``unknown unknowns'' that cannot be quantified at the time of an initial decision. These phenomena are relatively common in conservation practice \cite{doak2008understanding,wintle2010allocating}, and can be related to biological, environmental or sociopolitical factors \cite{schuster2022protected}. Potential research avenues to help better account for such phenomena include designing adaptive management programs that can simultaneously build general knowledge of a system while informing more specific decisions, and explicitly including risk-tolerance scenarios into management prescriptions.

\subsection{Causality: AI and economics to understand conservation impacts}
\label{sec:causality}

\subsubsection{The role of causal inference in conservation management}
Applied challenges in conservation science and management often raise difficult questions of cause and effect. \textit{Will additional anti-poaching patrols allow a threatened species to recover? What impact will a new conservation area have on deforestation?} Computer science, economics, and conservation biology have all made important advances in developing causal inference methods that enable researchers to use observational data to estimate causal effects \cite{butsic2017quasi,ferraro2019causal,pearl2009causal}. AI can improve our understanding of these causal relationships in multiple ways by, for example, (a)~extracting more useful information about the world from existing data; (b)~improving predictions of counterfactual scenarios of what might have happened in the absence of some intervention; or (c)~providing more data-driven approaches to evaluate heterogeneity in treatment effects \cite{storm2020machine}. 

However, data limitations often inhibit the direct use of off-the-shelf techniques from causal inference to study conservation impacts. Ecology settings commonly involve continuous real-time data, distribution shifts, and nonstationarity --- settings that are all challenging to study because ground truth is not easily available for evaluating the models. AI offers opportunities to expand causal studies by generating data on outcomes and improving causal estimates. 

While causal AI has made important contributions to all of these areas of research, our workshop's discussions primarily focused upon the \emph{potential} for AI to extract new information from data, as well as the \emph{methodological considerations} that these data necessitate for the goal of accuracy of causal estimates.

\paragraph{Potential to uncover causal relationships in conservation} Many rigorous research designs for observational, causal inference require consistent data spanning large geographic areas, measured over multiple periods of time. Given these needs, AI-based Earth observation has proven to be a valuable source of information for causal inference in conservation. 

However, these requirements have led the environmental economics and conservation biology communities to often focus on outcomes that are relatively easily measured. Rather than track a policy's impacts on the primary outcomes conservation practitioners truly care about (e.g., biodiversity and species populations), impact studies often evaluate changes in more easily measured proxies such as land cover and land use change. As we develop longer time series and broader spatial coverage of biodiversity data (based on e-DNA, audio sensors, and camera traps; see \cref{sec:earth-observation}), there is growing opportunity to more directly estimate treatment effects on the real variables of interest.

\paragraph{Methodological considerations for causal inference for conservation} The economics community often uses predictions from machine learning models as inputs into causal inference analyses, where an assumption might be made that derived data generated through AI can be integrated into existing causal inference research designs without significantly biasing estimated treatment effects. However, a growing body of research highlights that the structure of these data, and their underlying error distributions, require more careful reconsideration of causal inference techniques in order to generate unbiased estimates. For example, commonly-used, two-way fixed effects models may yield biased estimates of conservation impact when applied to pixel-level data detailing deforestation \cite{garcia2022conservation}. 

Careful consideration of the underlying data generating processes can help scientists structure their causal inference models to yield more accurate estimates of the impacts of conservation policies. More broadly, remotely sensed measures of environmental conditions exhibit both random and more structured measurement errors, which can bias estimates of causal relationships. In these settings, multiple imputation methods can generate useful error corrections to improve the accuracy of causal estimates \cite{proctor2023parameter}. This growing literature highlights the value of greater collaborations across disciplines --- a broader understanding of both the ways that machine learning data are generated and used can help identify ways to more reliably estimate the causal impacts of conservation interventions.

\paragraph{Overcoming missing data} Often in conservation, the outcomes managers and decision-makers care about are not directly observed in the data. Such missing data problems are studied in machine learning and statistics, including in areas such as positive-unlabeled learning. Conversely, statistical methods originally developed in ecology, such as missing species estimation, can be applied to other settings such as understanding where we might be missing data about urban incidents (fallen trees and sidewalk repairs) for responders to address \cite{liu2022equity}. These novel methods enable us to infer outcomes where there is no labeled outcome data.

\subsubsection{Open challenges in causality for conservation decision making}
In addition to the challenges of using predicted or remotely sensed variables in causal inference, we also highlight questions of external validity that arise in any causal inference. In particular, the following key challenges should come into consideration when studying causal relationships of conservation factors:

\paragraph{Shifts over time} A correctly identified causal relationship that holds in past data may or may not hold in current or future data. This presents an opportunity to design experiments, data collection protocols, and algorithms that integrate monitoring and optimization, trading off improving model fidelity with taking impactful intervention action.

\paragraph{Shifts over space} Trends identified in case studies may not generalize to other areas. This presents an opportunity to clearly communicate regions where the inference is and is not expected to be valid, and to design frameworks that increase the spatial extent that causal analysis can be applied to. 

\paragraph{Generalizability} How do we go from internal validity to external validity? This depends in part on the data we choose to make inferences on. Case studies may not generalize to other areas. It is essential to collect more datasets for distribution shifts in ecology settings. Creating systems to share data across geographic areas and time is a promising way to shed light on these challenges and provide a foundation for future methodological work.

\clearpage
\section{Critical considerations in AI-assisted decision making}
\label{sec:critical-considerations}

During our two-day workshop, several cross-cutting themes emerged throughout our sessions, highlighting critical considerations for AI-assisted decision making in conservation settings. In this section we synthesize those themes, grouped into: opportunities, limitations, and costs associated with using AI for conservation decision making. 

\subsection{Opportunities: What is AI-able?}
Contexts where data are plentiful, such as remote sensing and camera traps, are amenable to machine learning. In more data sparse situations, additional methods may be needed, for example synthetic or model-derived data. AI has found limited use in predictive regimes in the context of conservation since the necessary data, both historical and ecological context-rich, were typically sparse, incomplete, or missing altogether. As those data become available (\cref{sec:earth-observation}) --- in conjunction with the increasing availability of both current and historic data on human activity, weather, natural disasters, habitat, and more --- there is an opportunity to develop AI approaches that can assist conservation efforts. Below we outline three considerations for assessing such opportunities.

\paragraph{Scale} Conservation involves multiple scales (global to local) and different data are available at different scales. At a local scale there are clear challenges in data availability relating to the difficulty of data acquisition and data privacy issues. In particular, in conservation there are good reasons that data cannot be made FAIR \cite{wilkinson2016fair}. These include risks around endangered species and data sovereignty issues at multiple levels including governmental restrictions on data sharing and restrictions on sharing Indigenous and local knowledge (see the CARE principles \cite{carroll2020care}).

\paragraph{Metrics} Another challenge arises around the differences between outcome metrics (what real-world outcomes are desired) and AI performance metrics --- what mathematical objectives the AI system is designed to optimize given the existing data --- in a given conservation context. For example, biodiversity has a multifaceted definition which ``includes diversity within species, between species and of ecosystems'' \cite{international2000iucn}. In addition to biodiversity, conservation often focuses on ecosystem services such as pollination, water quality, carbon sequestration. And, for all aspects, benefits to people must be considered, including the health of communities on land being conserved or restored. Finally, the performance metrics used to evaluate the efficacy of AI approaches, such as species identification accuracy, are not necessarily well-aligned with outcome metrics. Performance metric improvement may or may not have a positive impact on the outcome of the entire conservation system, and this relationship is understudied.

\paragraph{Adaptability} Conservation often occurs in non-stationary situations, necessitating AI solutions that can adapt under shifting conditions. AI methods frequently see large performance losses when used \textit{out of distribution} --- i.e., in a scenario that doesn't match the data they were trained on \cite{koh2021wilds,beery2018recognition}. Understanding when and how much to trust the outcomes of an AI system in a changing world, and how to efficiently and cost effectively improve performance out of distribution is an open and impactful area of AI research paramount to the effectiveness of AI systems in conservation settings.

\subsection{Limitations: What is not AI-able?}
AI for conservation has the potential to address the challenge of scaling limited resources to significantly increase the area of well managed protected areas around the globe. For example, AI can help provide better predictive modeling capabilities than traditional spatial statistics. On the other hand, AI for conservation has the risk of promoting solutions that fail to scale, are more complex and costly than is necessary for the task, and may distract conservation agency staff from applying more appropriate decision-making techniques. 

As a community, we have a collective responsibility to identify solutions that will scale and have conservation impact, to be clear on our respective responsibilities and commitments, and to marshall the necessary resources to ensure these solutions are delivered. Conservation professions need guidance to allow them to better evaluate how to engage with conservation AI projects. Conversely, AI researchers and developers must invest in understanding practical challenges of conservation practitioners.

We recognize that while tech is often less than 20\% of the overall solution, computation and AI have a potential role to play in every part of the conservation monitoring lifecycle. We emphasize that AI decision-making methods should be adopted as decision aids, but cannot replace the expertise of human practitioners. One way to frame this is to design AI to help invoke critical thinking --- that is, AI to help enable decisions that then require conversations between human decision-makers to question the status quo of conservation initiatives, including defining what is beneficial in context, and potentially seeking alternative solutions to AI \cite{bondi2021envisioning}. 

\subsection{Costs: What are the costs of using AI?}
There are several costs associated with the choice to use AI as a part of a management system that may not be readily apparent. We saw that across many of our talks, the hidden costs of AI were a large factor contributing to the potential impact or reduction of impact. 

One factor that was brought up across many case studies was that the AI model was only 10\% of the work of integrating AI in conservation management frameworks. Agreeing on where the responsibility lies for the remaining 90\% of the work can be challenging, as there are differences of values and resources between academia, conservation land managers and organizations, and potential funding agencies or organizations. Further, in reality, AI will be only a component of a participatory system, as opposed to a fully-automated solution, and the use of AI will have ongoing computational and human costs.

We posit that it is important to assess all of the costs of AI use --- not just the costs of data collection and model training --- in deliberating the potential benefits of using of AI in conservation contexts. One key (and currently missing) component is low-cost assessment of AI feasibility for a new project, particularly in the case where there are existing off-the-shelf models that may be used. Making it cheap and easy to ask the question \textit{can I use this model for my project, and will it save me time and/or money?}\ should be a key priority. For example, the satellite-image embedding and prediction model in MOSAIKS \cite{rolf2021generalizable} prioritize model efficiency, accessibility, and speed of retraining to make this type of analysis lightweight. Projects like MOSAIKS evidence the value of re-evaluating AI performance metrics beyond accuracy to include usability metrics: \textit{How expensive is the model to train (including hyperparameter tuning)? How expensive is the model to use? How do we help users determine model uncertainty? In what percentage of the time do humans need to verify the results to have a system that is reliable enough for use?} 

\subsubsection{Making the costs of AI transparent}
Researchers and practitioners interested in applying AI to conservation decision making must account for the hidden costs of AI in deciding what to integrate into their practices. One important step forward is making the full costs of developing and maintaining AI solutions more transparent from the outset.  There are hidden costs across the entire pipeline of AI development:

\paragraph{Data collection and analysis} Data collection is costly and includes the purchase of sensors or data, acquisition of labels, management of the data, storage of the data, and accessibility and standardization of the data. Additionally, cost of preliminary data analysis for ML readiness should be accounted for, including data science capacity cost and computational capacity cost.

\paragraph{Model development} We must consider the cost of ML prototyping. This can be significantly more expensive if a new question is being asked and no existing labeled datasets, best practices for model training, or pretrained models are available for this task. Costs to be considered include GPU costs of training, which increase with data size, model size, and model robustness to hyperparameters. Note also that reinforcement learning and meta learning are much more expensive computationally than training a classification model. Smaller machine learning architectures, such as ResNet, are much cheaper to train and evaluate from both a size and robustness perspective than larger architectures. 

\paragraph{System infrastructure} Costly data infrastructure development is often necessary to scale the use of ML beyond a prototype. \textit{How will the data get from your collection sites to the model? Where will the data be stored (locally or on the cloud), where will the model be stored, what are the costs of data movement, and what temporal latency do you need for your problem?} Larger models are also more costly to use for inference, either on the cloud or through the purchase of additional GPUs, and this increased cost should be weighed against performance gains that may be achieved with larger models. 

\paragraph{Evaluation} We must both plan for and account for the costs of persistent quality control. \textit{How will a team verify and analyze performance of the model to assess bias? If the model is no longer working well, what will you do? Do you need to retrain your model periodically?} If this is the case, we must carefully consider what data and decision making infrastructure are needed to retrain models, and account for these costs in our planning.

\paragraph{Human resources} Successful impact-driven, interdisciplinary collaboration requires extensive communication of both expectations and results. Human expertise ought to be integrated in the development of AI models \cite{kuhnert2010guide}. If an AI partner pulls out of a collaboration before delivering a usable conservation technology tool to the end beneficiary, beyond breaching of trust, there is a reputational cost for other AI researchers. \textit{If one of the players in the development of these models pulls out, what is the sunk cost for the other contributors and is that cost recoverable?} For example, if a conservation organization sinks considerable resources into ML data curation and then a usable model cannot be developed, \textit{is that effort aligned with better data infrastructure and management for that organization in a way that is useful even without ML?}

\paragraph{Environmental} Finally, all of the above costs are both financial and environmental. Both computational resources and data storage use power, which has an associated carbon and natural resource cost. We must be aware of the carbon and nature costs of the systems we build and account for these in our metrics and in our decision-making processes \cite{schwartz2020green}.

\subsection{Guiding question: How to design AI that aids humans in conservation decision making?}
Our talks and discussions illuminated a number of recurring themes on the ways in which the demands of human decision making require solutions that move beyond simply producing predictions or even suggesting future actions: 
\begin{enumerate}
    \item The need for understandable, computationally inexpensive, user-friendly, and easily deployed procedures. 
    \item The need for quantitative assessments of actual downstream conservation impact to judge the effectiveness of AI tools and interventions. 
    \item The need to determine which stakeholders or rights holders make which decisions and what form of information they need to make those decisions \cite{moreto2015poaching}. 
    \item The importance of accounting for financial, social, and human effort costs in selecting which AI tools to deploy. 
    \item Model interpretability arose as an important factor for model usability and value in the hands of stakeholders \cite{kuiper2020rangers}. 
    \item Model bias, fairness, and trustworthiness all impact long term user adoption and satisfaction. 
\end{enumerate}
Each of these motifs in isolation highlights important questions, but, importantly, AI for conservation requires the \emph{intersection} of these considerations. Moreover, important cultural changes are needed in the AI discipline to incentivize and reward the work needed to address these challenges.

AI must aid decision making in conservation, not replace the expert decision process with fully-automated systems. Due to the high risk associated with errors that AI for conservation systems often face, similar to AI for healthcare or self driving cars, there is a need for additional research and development of participatory systems that enable human-AI collaboration \cite{bondi2022role,norouzzadeh2021deep,miao2021iterative}. These systems and methods seek not to replace human expertise but instead to augment it, making efficient use of human intelligence and intervention and enabling a single expert to process far more --- and far more complex --- data than was previously possible. 

One example of such a system is ElephantBook \cite{kulits2021elephantbook}, a system which helps a small group of experts in the field monitor an elephant population of over 2,000 individuals by speeding up the visual re-identification process. Rangers and local experts photograph elephants when they are sighted, and the system combines expert individual codes with modern computer vision methods to find the best possible matches in the database of local known individuals (which includes over 50K images). Human experts sort through the possible matches and verify them, making the most of both machine efficiency and human ability to generalize and recognize patterns from only a few examples. This again points to a need to broaden  our range of AI evaluation metrics, as a successful system should jointly meet accuracy goals needed for the task while minimizing the use of human time and resources.

Finally, one major remaining challenge is knowing when and how much to trust AI predictions or recommendations, particularly given that these models often have the most \emph{potential} for added value when deployed outside of the distribution on which they were trained. We need to invest research cycles on determining both how best to estimate uncertainty and how to effectively communicate those uncertainty estimates. Here we can learn from adjacent domains: these types of uncertainty estimates and how they are best communicated have already been explored in weather forecasts \cite{morss2008communicating} and in climate models \cite{huntingford2019machine}.

\clearpage
\section{Looking forward to the future of AI-assisted decision making for conservation}

The 2022 AI-Assisted Decision Making for Conservation workshop convened to explore opportunities to expand the breadth of AI methods --- particularly for optimization, planning, and evaluation --- that are being developed for and applied to conservation. Our participants came from across academia, industry, and non-profits with expertise in computation, conservation, and deploying research. Together, we shared our expertise and hopes for the field of AI-assisted decision making in conservation, as it progresses.\footnote{Some ideas and goals were shared by individuals, some in small groups, and some were common to all participants in our group. Here we summarize and synthesize these points without necessarily differentiating between these cases.} 

We end by highlighting forward-looking considerations on the interconnectedness of different conservation management goals in \ref{sec:conclusion_intersectionality} and summarizing prominent goals and ideas for moving forward in section \ref{sec:vision_ideas}.

\subsection{Intersectionality of conservation management goals and its implications for useful AI}
\label{sec:conclusion_intersectionality}
In discussing AI for conservation, it is important to simultaneously consider social and financial concerns rather than purely focusing on just environmental goals.  Environmental goals should be socially viable. As a specific example, success in wildlife conservation efforts may lead to increased human-wildlife conflicts \cite{dickman2010complexities}; such costs must be considered and mitigated to the extent possible. 

Similarly, environmental goals should be financially viable. Sustainable solutions in AI for conservation would allow us to simultaneously address all three concerns (see \cref{fig:sustainability-venn}). Achieving these simultaneous goals requires deep partnership with local communities, non-profits, governments, and cannot be achieved by AI researchers in isolation, armed only with datasets. 

It is unrealistic to expect to be able to assess all intersectional costs and interactions beforehand. Rather, we might develop and rely upon key partnerships to continually reassess the impact of the AI deployment. A staged approach to deployment with continuous assessment may help mitigate or address emerging financial or social challenges.

\subsection{A vision for the future directions in AI-assisted conservation}
\label{sec:vision_ideas}
To summarize, AI is being used widely for conservation \cite{tuia2022perspectives,gomes2019computational,bakker2022smart}, and holds promise for even more uses across a vast and intersectional range of conservation efforts. As we expand new frontiers of AI for conservation, it is useful to orient this vision around the following questions: 
\begin{itemize}
    \item What questions in conservation are answerable and what decisions are supportable with AI? Including:
    \begin{itemize}
        \item When and how can AI approaches effectively assist conservation decision making and prioritization?
        \item When and how can AI help assess conservation impacts and outcomes?
    \end{itemize}
    \item What are the full costs associated with developing and deploying AI systems for conservation? How do these compare to the intended benefits?
\end{itemize}

\vspace{8ex}
\subsection*{Acknowledgments}

We thank Carl Boettiger (UC Berkeley), Iadine Chad\'{e}s (CSIRO), Ross Tyzack Pitman (Panthera), and Michael C. Runge (USGS) for helpful discussions prior to the workshop. 

This workshop was funded by the Center for Research on Computation and Society at Harvard University. Thank you to Emma Hammack and Hila Bernstein of CRCS for all your support!

\clearpage
\small

\bibliographystyle{unsrtnat}
\bibliography{references}

\end{document}